\documentclass[10pt,twocolumn,letterpaper]{article}

\usepackage{cvpr}
\usepackage{times}
\usepackage{epsfig}
\usepackage{graphicx}
\usepackage{amsmath}
\usepackage{amssymb}

\usepackage{textcomp}
\usepackage{svg}
\usepackage{bm}

\usepackage[breaklinks=true,bookmarks=false]{hyperref}

\cvprfinalcopy %

\def\cvprPaperID{6924} %

\ifcvprfinal\fi
\begin{document}

\title{Weakly-Supervised Action Localization
by Generative Attention Modeling}

\author{Baifeng Shi$^{1}$\thanks{Work was done during internship at Microsoft.}~~~~~Qi Dai$^{2}$~~~~~Yadong Mu$^{1}$~~~~Jingdong Wang$^{2}$\\
  $^1$Peking University~~~~~
  $^2$Microsoft Research Asia\\
  {\tt\small \{bfshi,myd\}@pku.edu.cn, \{qid,jingdw\}@microsoft.com}
}

\maketitle

\begin{abstract}
Weakly-supervised temporal action localization is a problem of learning an action localization model with only video-level action labeling available.
The general framework largely relies on the classification activation, which employs an attention model to identify the action-related frames and then categorizes them into different classes.
Such method results in the action-context confusion issue: context frames near action clips tend to be recognized as action frames themselves, since they are closely related to the specific classes.
To solve the problem, in this paper we propose to model the class-agnostic frame-wise probability conditioned on the frame attention using conditional Variational Auto-Encoder (VAE).
With the observation that the context exhibits notable difference from the action at representation level, a probabilistic model, i.e., conditional VAE, is learned to model the likelihood of each frame given the attention.
By maximizing the conditional probability with respect to the attention, the action and non-action frames are well separated.
Experiments on THUMOS14 and ActivityNet1.2 demonstrate advantage of our method and effectiveness in handling action-context confusion problem. Code is now available on GitHub\footnote{https://github.com/bfshi/DGAM-Weakly-Supervised-Action-Localization}.

\end{abstract}

\section{Introduction}
\label{sec1}
Action localization is one of the most challenging tasks in video analytics and understanding~\cite{simonyan2014two,shou2016temporal,li2018recurrent,qiu2019learning,li2019long}.
The goal is to predict accurate start and end time stamps of different human actions.
Owing to its wide application (\eg, surveillance~\cite{vishwakarma2013survey}, video summarization~\cite{hua2005generic}, highlight detection~\cite{xiong2019less}), action localization has drawn lots of attention in the community.
Thanks to the powerful convolutional neural network (CNN)~\cite{krizhevsky2012imagenet}, performance achieved on this task has gone through a phenomenal surge in the past few years \cite{shou2016temporal,yeung2016end,dai2017temporal,xu2017r,chao2018rethinking,alwassel2018action,lin2018bsn,long2019gaussian}.
Nevertheless, these fully-supervised methods require temporal annotations of action intervals during training, which is extremely expensive and time-consuming.
Therefore, the task of weakly-supervised action localization (WSAL) has been put forward, where only video-level category labels are available.

To date in the literature, there are two main categories of approaches in WSAL.
The first type~\cite{liu2019completeness,narayan20193c,paul2018w,wang2017untrimmednets} generally builds a top-down pipeline, which learns a video-level classifier and then obtains frame attention by checking the produced temporal class activation map (TCAM)~\cite{zhou2016learning}.
Note that a \emph{frame} indicates a small snippet from which appearance or motion feature could be extracted.
On the other hand, the second paradigm works in a bottom-up way, \ie, temporal attention is directly predicted from raw data~\cite{nguyen2018weakly,nguyen2019weakly,shou2018autoloc,yuan2018marginalized}. Then attention is optimized in the task of video classification with video-level supervision.
Frames with high attention are thus treated as action part, otherwise the background part.

\begin{figure}
    \centering
    \includegraphics[width=1\linewidth]{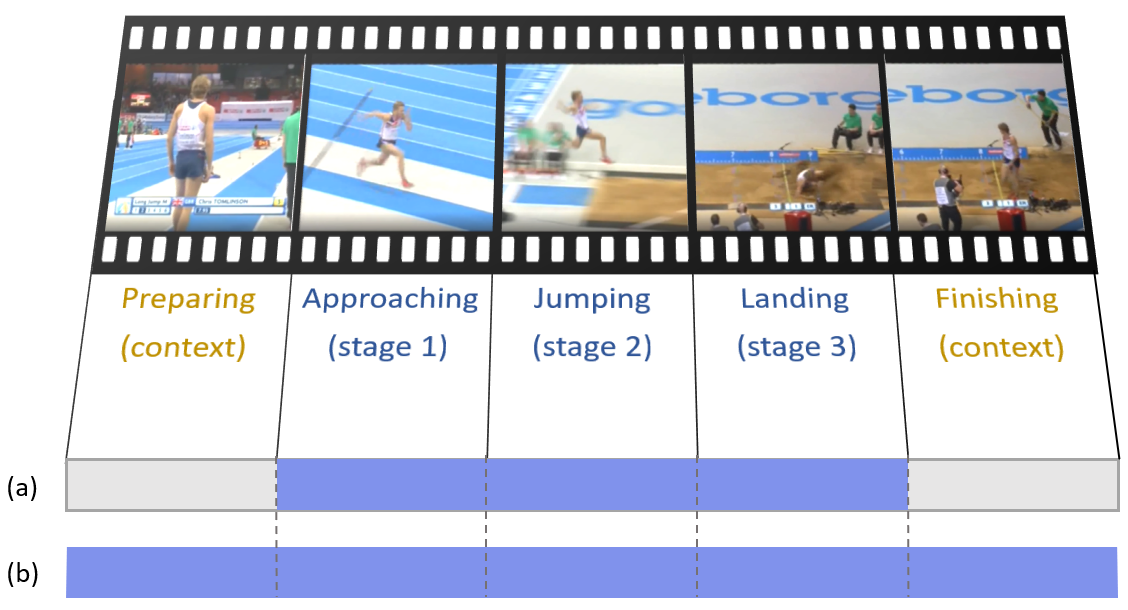}
    \caption{An illustration of action-context confusion. The video clip, showing a \emph{long jump} process, consists of three stages of the action (approaching, jumping, and landing) and two stages of context (preparing and finishing). (a) Ground truth of action localization. (b) Action-context confusion. The context frames, which are highly related to the \emph{long jump} category, are also selected.}
    \label{problem}
\end{figure}

Both kinds of methods largely rely on the video-level classification model, which would lead to the intractable action-context confusion~\cite{liu2019completeness} issue in the absence of frame-wise labels.
Take the \emph{long jump} in Figure \ref{problem} as an example, the action has three stages, \ie, approaching, jumping, and landing.
In addition, the frames before and after the action, \ie, preparing and finishing, contain the content that is closely related to \emph{long jump}, but are not parts of the action.
We refer to such frames as context, which is a special kind of background.
In this example, the context parts include the track field and sandpit, which could in fact significantly encourage the recognition of the action.
Without frame-wise annotations, the classifier is normally learned by aggregating the features of all related frames, where context and action are roughly mixed up.
The context frames thus tend to be easily recognized as action frames themselves.
The action-context confusion problem has not been fully studied though it is common in WSAL.
One recent exploration \cite{liu2019completeness} attempts to solve the problem by assuming a strong prior that context clips should be stationary, \ie, no motions in them.
However, such assumption has massive limitations and ignores the inherent difference between context and action.

To separate context and action, the model should be able to capture the underlying discrepancy between them.
Intuitively, context frame indeed exhibits obvious difference from action frame at the appearance or motion level.
For example, among the five stages in Figure \ref{problem}, the action stages (approaching, jumping, and landing) clearly demonstrate more intense body postures than the context stages (preparing and finishing).
In other words, the extracted feature representations for context and action are also different.
Such difference exists regardless of the action category.

Inspired by this observation, we propose a novel generative attention mechanism to model the frame representation conditioned on frame attention. %
In addition to the above intuition, we build a graphical model to theoretically demonstrate that the localization problem is associated with both the conventional classification and the proposed representation modeling.
Our framework thus consists of two parts: the Discriminative and Generative Attention Modeling (DGAM).
On one hand, the discriminative attention modeling trains a classification model on temporally pooled features weighted by the frame attention.
On the other hand, a generative model, \ie, conditional Variational Auto-Encoder (VAE), is learned to model the class-agnostic frame-wise distribution of representation conditioned on attention values.
By maximizing likelihood of the representation, the frame-wise attention is optimized accordingly, leading to well separation of action and context frames.
Extensive experiments are conducted on THUMOS14~\cite{idrees2017thumos} and ActivityNet1.2~\cite{caba2015activitynet} to show that DGAM outperforms the state-of-the-arts by a significant margin.
Comprehensive analysis further validates its effectiveness on separating action and context.

The main contribution of this work is the proposed DGAM framework for addressing the issue of action-context confusion in WSAL by modeling the frame representation conditioned on different attentions.
The solution has led to elegant views of how localization is associated with the representation distribution
and how to learn better attentions by modeling the representation,
which have not been discussed in the literature.

\section{Related Works}

\textbf{Video action recognition} is a fundamental problem in video analytics.
Most video-related tasks leverage the off-the-shelf action recognition models to extract features for further analysis.
Early methods normally devise hand-crafted features~\cite{laptev2005space,wang2013action,oneata2013action} for recognition.
Recently, thanks to the development of deep learning techniques, lots of approaches focus on automatic feature extraction with end-to-end learning, \eg, two-stream network~\cite{simonyan2014two}, temporal segment network (TSN)~\cite{wang2016temporal}, 3D ConvNet (C3D)~\cite{tran2015learning}, Pseudo 3D (P3D)~\cite{qiu2017learning}, Inflated 3D (I3D)~\cite{carreira2017quo}.
In our experiments, I3D is utilized for feature extraction.

\textbf{Fully-supervised action localization} has been extensively studied recently.
Many works follow the paradigms that are widely applied in object detection area~\cite{girshick2014rich,girshick2015fast,ren2015faster,redmon2016you,liu2016ssd} due to their commonalities in problem setting.
To be more specific, there are mainly two directions, namely two-stage method and one-stage method.
Two-stage methods~\cite{zhao2017temporal,xu2017r,dai2017temporal,chao2018rethinking,shou2016temporal,shou2017cdc,gao2017cascaded,heilbron2017scc,lin2018bsn} first generate action proposals and then classify them with further refinement on temporal boundaries.
One-stage methods~\cite{buch2017end,lin2017single,zhang2018s3d} instead predict action category and location directly from raw data.
In fully-supervised setting, the action-context confusion could be alleviated with frame-wise annotations.

\textbf{Weakly-supervised action localization} is drawing increasing attention due to the time-consuming manual labeling in fully-supervised setting.
As introduced in Section \ref{sec1}, WSAL methods can be grouped into two categories, namely top-down and bottom-up methods.
In top-down pipeline (\eg UntrimmedNet~\cite{wang2017untrimmednets}), video-level classification model is learned first, and then frames with high classification activation are selected as action locations.
W-TALC~\cite{paul2018w} and 3C-Net~\cite{narayan20193c} also force foreground features from the same class to be similar, otherwise dissimilar.
Unlike top-down scheme, the bottom-up methods directly produce the attention for each frame from data, and train a classification model with the features weighted by attention.
Based on this paradigm, STPN~\cite{nguyen2018weakly} further adds a regularization term to encourage the sparsity of action.
AutoLoc~\cite{shou2018autoloc} proposes the Outer-Inner-Contrastive (OIC) loss by assuming that a complete action clip should look different from its neighbours.
MAAN~\cite{yuan2018marginalized} proposes to suppress dominance of the most salient action frames and retrieve less salient ones.
Nguyen \etal~\cite{nguyen2019weakly} propose to penalize the discriminative capacity of background, which is also utilized in our classification module.
Besides, a video-level clustering loss is applied in~\cite{nguyen2019weakly} to separate foreground and background.
Nevertheless, all of the aforementioned methods ignore the challenging action-context confusion issue caused by the absence of frame-wise label.
Though Liu \etal~\cite{liu2019completeness} try to separate action and context using hard negative mining, their method is based on the strong assumption that context clips should be stationary, which has many limitations and may hence cause negative influence on the prediction.

\textbf{Generative model} has also experienced a fast development in recent years \cite{kingma2013auto,goodfellow2014generative,higgins2017beta}.
GAN~\cite{goodfellow2014generative} employs a generator to approximate real data distribution by the adversarial training between generator and discriminator.
However, the learned approximating distribution is implicitly determined by generator and thus cannot be analytically expressed.
VAE~\cite{kingma2013auto} approximates the real distribution by optimizing the variational lower bound on the marginal likelihood of data.
Given a latent code, the conditional distribution is explicitly modeled as a Gaussian distribution, hence data distribution can be analytically expressed by sampling latent vectors and calculating the Gaussian. %
Flow-based model~\cite{kingma2018glow} uses invertible layers as the generative mapping, where data distribution can be calculated given the Jacobian of each layer.
However, all layers must have the same dimensions, which is much less flexible.
In our work, we exploit Conditional VAE (CVAE)~\cite{sohn2015learning} to model the frame feature distribution conditioned on attention value.

\section{Method}

Suppose we have a set of training videos and the corresponding video-level labels.
For each video, we sample $T$ frames (snippets) to extract the RGB or optical flow features $\mathbf{X} = (\mathbf{x}_t)_{t = 1}^{T}$ with a pre-trained model, where $\mathbf{x}_t \in \mathbb{R}^d$ is the feature of frame $t$, and $d$ is feature dimension.
The video-level label is denoted as $y \in \{0, 1, \cdots, C\}$, where $C$ is the number of classes and $0$ corresponds to background.
For brevity, we assume that each video only belongs to one class, though the following discussion can also apply to multi-label videos.

Our method follows the bottom-up pipeline for WSAL, which learns the attention $\bm{\lambda} = (\lambda_t)_{t = 1}^{T}$ directly from data, where $\lambda_t \in [0, 1]$ is the attention of frame $t$.
Before discussing the details of our method, we examine the action localization problem from the beginning.

\subsection{Attention-based Framework}

In attention-based action localization problem, the target is to predict the frame attention $\bm{\lambda}$, which is equivalent to solving the maximum a posteriori (MAP) problem:
\begin{equation}\small
\label{OP1}
    \max_{\lambda_t \in [0, 1]} \log p(\bm{\lambda} | \mathbf{X}, y),
\end{equation}
where $p(\bm{\lambda} | \mathbf{X}, y)$ is the unknown probability distribution of $\bm{\lambda}$ given $\mathbf{X}$ and $y$.
In the absence of frame-level labels (ground truth of $\bm{\lambda}$), it is difficult to approximate and optimize $p(\bm{\lambda}|\mathbf{X}, y)$ directly.
Therefore, we transform the optimization target using Bayes\textquotesingle~theorem,
\begin{equation}\small
    \begin{split}
        \log p(\bm{\lambda} | \mathbf{X}, y) &= \log p(\mathbf{X}, y | \bm{\lambda}) + \log p(\bm{\lambda}) - \log p(\mathbf{X}, y) \\
        &= \log p(y | \mathbf{X}, \bm{\lambda}) + \log p(\mathbf{X} | \bm{\lambda}) + \log p(\bm{\lambda}) \\ & \quad - \log p(\mathbf{X}, y) \\
        &\propto \log p(y | \mathbf{X}, \bm{\lambda}) + \log p(\mathbf{X} | \bm{\lambda}),
    \end{split}
\end{equation}
where in the last step, we discard the constant term $\log p(\mathbf{X}, y)$ and assume a uniform prior of $\bm{\lambda}$, \ie, $p(\bm{\lambda}) = const$. Our optimization problem thus becomes
\begin{equation}\small
\label{OP2}
    \max_{\lambda \in [0, 1]} \log p(y | \mathbf{X}, \bm{\lambda}) + \log p(\mathbf{X} | \bm{\lambda}).
\end{equation}
This formulation indicates two different aspects for optimizing $\bm{\lambda}$.
The first term $\log p(y | \mathbf{X}, \bm{\lambda})$ prefers $\bm{\lambda}$ with high \textbf{discriminative} capacity for action classification, which is the main optimization target in previous works.
In contrast, the second term $\log p(\mathbf{X} | \bm{\lambda})$ forces the representation of frames to be accurately predicted from the attention $\bm{\lambda}$.
Given the feature difference between foreground and background, this objective encourages the model to impose different attentions on different features.
In specific, we exploit a \textbf{generative} model to approximate $p(\mathbf{X} | \bm{\lambda})$, and force the feature $\mathbf{X}$ to be accurately reconstructed by the model.

\begin{figure}[t]
  \centering
  \includegraphics[width=0.5\linewidth]{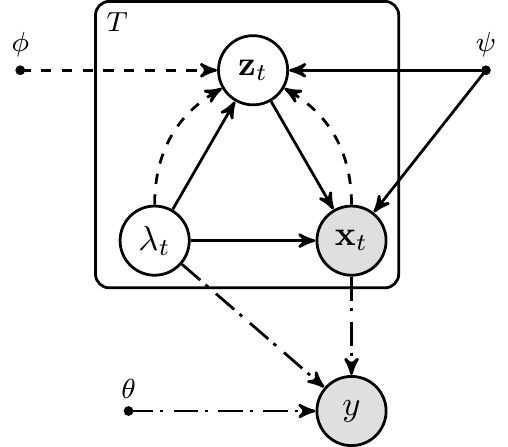}
  \caption{The directed graphical model of DGAM. Solid lines denote the generative model $p_{\psi}(\mathbf{z}_t | \lambda_t) \ p_{\psi}(\mathbf{x}_t | \lambda_t,\mathbf{z}_t)$, dashed lines denote the variational approximation $q_{\phi}(\mathbf{z}_t | \mathbf{x}_t,\lambda_t)$ to intractable posterior $p(\mathbf{z}_t | \mathbf{x}_t,\lambda_t)$, and dash-dot lines denote the video-level classification model $p_{\theta}(y | \mathbf{x}_t,\lambda_t)$. $\phi$ and $\psi$ are jointly learned, which forms an alternating optimization together with $\theta$ and $\lambda_t$. }
\label{graphicModel}
\end{figure}

\begin{figure*}[htbp]
  \centering
  \includegraphics[width=1\linewidth]{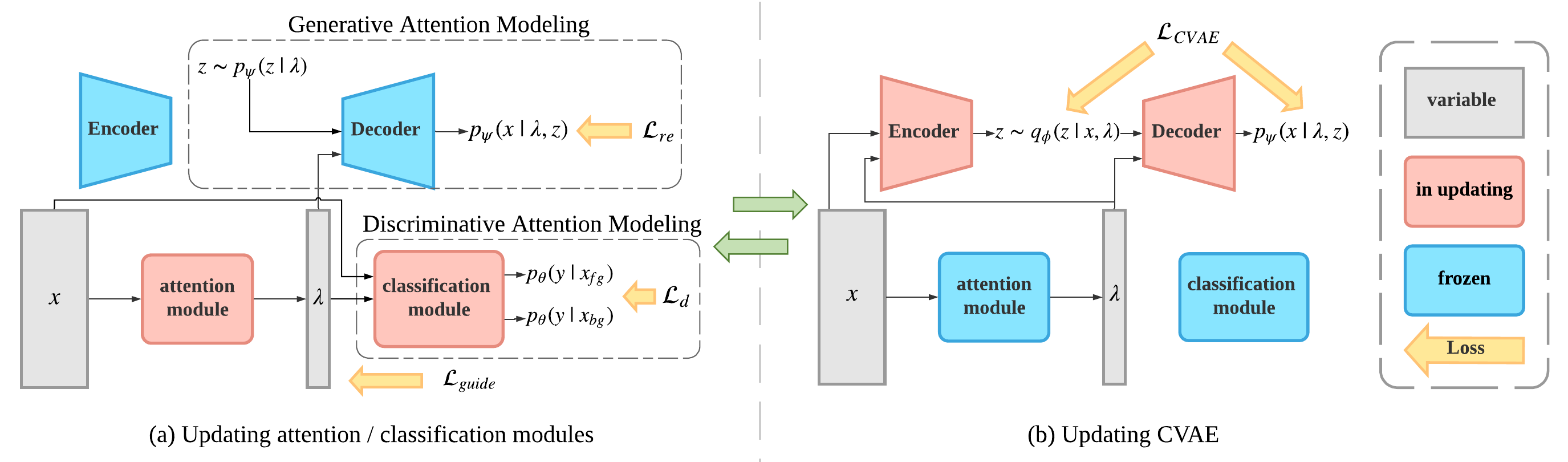}
  \caption{\textbf{Framework overview.} The proposed model is trained in two alternating stages (a) and (b).
  In stage (a), the generative model (CVAE) is frozen. Attention module and classification module are updated with classification-based discriminative loss $\mathcal{L}_d$, representation-based reconstruction loss $\mathcal{L}_{re}$ and regularization loss $\mathcal{L}_{guide}$. In stage (b), attention and classification modules are frozen. The CVAE is trained with loss $\mathcal{L}_{CVAE}$ to reconstruct the representation of frames with different $\lambda$. Since the ground truth $\lambda$ is unavailable, we utilize $\lambda$ predicted by attention module as ``pseudo label'' for training. }
\label{overall architecture}
\end{figure*}

Figure \ref{graphicModel} shows the graphical model of the above problem. The model parameters ($\theta,\psi,\phi$) and the latent variables in generative model ($\mathbf{z}_t$) will be discussed later.
Based on (\ref{OP2}), the framework of our method consists of two components, \ie, the discriminative attention modeling and the generative attention modeling, as illustrated in Figure \ref{overall architecture}.

\subsection{Discriminative Attention Modeling}

The discriminative attention module learns the frame attention by optimizing the video-level recognition task.
In specific, we utilize attention $\bm{\lambda}$ as weight to perform temporal average pooling over all frames in the video and produce a video-level \emph{foreground} feature $\mathbf{x}_{fg} \in \mathbb{R}^d$ given by
\begin{equation}\small
    \mathbf{x}_{fg} = \frac{\sum_{t = 1}^{T} \lambda_t \mathbf{x}_t}{\sum_{t = 1}^{T} \lambda_t}.
\end{equation}
Similarly, we can also utilize $\mathbf{1} - \bm{\lambda}$ as the weight to calculate a \emph{background} feature $\mathbf{x}_{bg}$:
\begin{equation}\small
    \mathbf{x}_{bg} = \frac{\sum_{t = 1}^{T} (1 - \lambda_t) \mathbf{x}_t}{\sum_{t = 1}^{T} (1 - \lambda_t)}.
\end{equation}
To optimize $\bm{\lambda}$, we encourage high discriminative capability of the foreground feature $\mathbf{x}_{fg}$ and simultaneously punish any discriminative capability of the background feature $\mathbf{x}_{bg}$~\cite{nguyen2019weakly}.
This is equivalent to minimizing the following discriminative loss (\ie softmax loss):
\begin{equation}\small
\label{L_d}
    \mathcal{L}_d = \mathcal{L}_{fg} + \alpha \cdot \mathcal{L}_{bg} =-\log p_\theta (y | \mathbf{x}_{fg}) - \alpha \cdot \log p_\theta (0 | \mathbf{x}_{bg}),
\end{equation}
where $\alpha$ is a hyper-parameter, and $p_\theta$ is our classification module modeled by a fully-connected layer with weight $\mathbf{w}_c \in \mathbb{R}^d$ for each class $c$ and a following softmax layer. During training, attention module and classification module are jointly optimized.
The graphical model of this part is illustrated in Figure \ref{graphicModel} with dash-dot lines.

\subsection{Generative Attention Modeling}

The discriminative attention optimization generally has difficulty in separating context and foreground when frame-wise annotations are unavailable.
Based on the observation that context differs from foreground in terms of feature representation, we utilize a Conditional Variational Auto-Encoder (CVAE) to model the representation distribution of different frames.
Before explaining the details, we briefly review the Variational Auto-Encoder (VAE).

Given the observed variable $\mathbf{x}$, \textbf{VAE} \cite{kingma2013auto} introduces a latent variable $\mathbf{z}$, and aims to generate $\mathbf{x}$ from $\mathbf{z}$, i.e.,
\begin{equation}
\label{equation_review_vae1}
    p_{\psi}(\mathbf{x})=\mathbb{E}_{p_{\psi}(\mathbf{z})} [p_{\psi}(\mathbf{x} | \mathbf{z})],
\end{equation}
where $\psi$ denotes the parameters of generative model, $p_{\psi}(\mathbf{z})$ is the prior (\eg a standard Gaussian),
and $p_{\psi}(\mathbf{x} | \mathbf{z})$ is the conditional distribution indicating the generation procedure, which is typically estimated with a neural network $f_{\psi}(\cdot)$ that is referred to as \emph{decoder}.
The key idea behind is to sample values of $\mathbf{z}$ that are likely to produce $\mathbf{x}$, which means that we need an approximation $q_{\phi}(\mathbf{z}|\mathbf{x})$ to the intractable posterior $p(\mathbf{z}|\mathbf{x})$.
$\phi$ denotes the parameters of approximation model, and $q_{\phi}(\mathbf{z}|\mathbf{x})$ is also estimated via a neural network $f_{\phi}(\cdot)$, which is referred to as \emph{encoder}.
VAE incorporates encoder $q_{\phi}(\mathbf{z}|\mathbf{x})$ and decoder $p_{\psi}(\mathbf{x} | \mathbf{z})$, and learns parameters by maximizing the variational lower~bound:
\begin{equation}\small
\begin{split}
\label{equation_review_vae2}
    \mathcal{J}_{VAE} &= -KL(q_{\phi}(\mathbf{z}|\mathbf{x})||p_{\psi}(\mathbf{z}))+\mathbb{E}_{q_{\phi}(\mathbf{z}|\mathbf{x})}[\log p_{\psi}(\mathbf{x} | \mathbf{z})],
\end{split}
\end{equation}
where $KL(q||p)$ is the KL divergence of $p$ from $q$.

In our DGAM model, we expect to generate the observation $\mathbf{X}$ based on the attention $\bm{\lambda}$, \ie, $p(\mathbf{X} | \bm{\lambda})$, which can be written as $p(\mathbf{X} | \bm{\lambda}) = \Pi_{t = 1}^{T} p(\mathbf{x}_t | \lambda_t)$ by assuming independence between frames in a video.
Similarly, we introduce a latent variable $\mathbf{z}_t$, and attempt to generate each $\mathbf{x}_t$ from $\mathbf{z}_t$ and $\lambda_t$, which forms a Conditional VAE problem:
\begin{equation}
\label{equation_cvae1}
    p_{\psi}(\mathbf{x}_t|\lambda_t)=\mathbb{E}_{p_{\psi}(\mathbf{z}_t|\lambda_t)} [p_{\psi}(\mathbf{x}_t |\lambda_t, \mathbf{z}_t)].
\end{equation}
Note that the desired distribution of $\mathbf{x}_t$ is modeled as a Gaussian, \ie, $p_{\psi}(\mathbf{x}_t |\lambda_t, \mathbf{z}_t)=\mathcal{N}(\mathbf{x}_t | f_{\psi}(\lambda_t,\mathbf{z}_t),\sigma^2*\mathbf{I})$, where $f_{\psi}(\cdot)$ is the decoder, $\sigma$ is a hyper-parameter, and $\mathbf{I}$ is the unit matrix.
Ideally, $\mathbf{z}_t$ is sampled from the prior $p_{\psi}(\mathbf{z}_t|\lambda_t)$.
In DGAM, we set the prior as a Gaussian, \ie, $p_\psi (\mathbf{z}_t | \lambda_t)=\mathcal{N}(\mathbf{z}_t|r \lambda_t \cdot \mathbf{1},~\mathbf{I})$, where $\mathbf{1}$ is all-ones vector and $r$ is a hyper-parameter indicating the discrepancy between priors of different attention value $\lambda_t$.
When $r = 0$, prior $p_\psi (\mathbf{z}_t | \lambda_t)$ is independent of $\lambda_t$.

During training of CVAE, we also approximate the intractable posterior $p(\mathbf{z}_t | \mathbf{x}_t, \lambda_t)$ by a Gaussian $q_\phi(\mathbf{z}_t | \mathbf{x}_t, \lambda_t)=\mathcal{N}(\mathbf{z}_t|\mu_{\phi},\Sigma_{\phi})$, where $\mu_{\phi}$ and $\Sigma_{\phi}$ are the outputs of the encoder $f_{\phi}(\mathbf{x}_t,\lambda_t)$.
We then minimize the variational loss $\mathcal{L}_{CVAE}$:
\begin{equation}\small
\begin{split}
\label{L_CVAE}
    \mathcal{L}_{CVAE} &= -\mathbb{E}_{q_\phi(\mathbf{z}_t | \mathbf{x}_t, \lambda_t)} \log p_\psi (\mathbf{x}_t | \lambda_t, \mathbf{z}_t) \\
    &\quad \ +  \beta \cdot KL(q_\phi (\mathbf{z}_t | \mathbf{x}_t, \lambda_t) || p_\psi (\mathbf{z}_t | \lambda_t))\\
    &\simeq -\frac{1}{L} \sum_{l = 1}^{L} \log p_\psi (\mathbf{x}_t | \lambda_t, \mathbf{z}_t^{(l)}) \\
    &\quad \ +  \beta \cdot KL(q_\phi (\mathbf{z}_t | \mathbf{x}_t, \lambda_t) || p_\psi (\mathbf{z}_t | \lambda_t)),
\end{split}
\end{equation}
where $\mathbf{z}_t^{(l)}$ is $l$-th sample from $q_\phi(\mathbf{z}_t | \mathbf{x}_t, \lambda_t)$.
Note that the Monte Carlo estimation of the expectation $\mathbb{E}_{q_\phi(\mathbf{z}_t | \mathbf{x}_t, \lambda_t)}(\cdot)$ is employed
with $L$ samples.
$\beta$ is a hyper-parameter for trade-off between reconstruction quality and sampling accuracy.%

For the generative attention modeling of $\bm{\lambda}$, we fix CVAE and minimize the reconstruction loss $L_{re}$ given by
\begin{equation}\small
\begin{split}
\label{eqn:L_re_1}
    \mathcal{L}_{re} %
    &= -\sum_{t = 1}^{T} \log \Big \{ \mathbb{E}_{p_\psi(\mathbf{z}_t | \lambda_t)} [p_\psi (\mathbf{x}_t | \lambda_t, \mathbf{z}_t)] \Big \} \\
    &\simeq -\sum_{t = 1}^{T} \log  \Big \{ \frac{1}{L} \sum_{l = 1}^{L} p_\psi (\mathbf{x}_t | \lambda_t, \mathbf{z}_t^{(l)}) \Big \},
\end{split}
\end{equation}
where $\mathbf{z}_t^{(l)}$ is sampled from the prior $p_{\psi}(\mathbf{z}_t|\lambda_t)$.
In our experiments, $L$ is set to $1$, and (\ref{eqn:L_re_1}) can be written as
\begin{equation}\small
\label{L_re}
    \mathcal{L}_{re} = -\sum_{t = 1}^{T} \log p_\psi (\mathbf{x}_t | \lambda_t, \mathbf{z}_t) \propto \sum_{t = 1}^{T} ||\mathbf{x}_t-f_{\psi}(\lambda_t,\mathbf{z}_t)||^2.
\end{equation}
The graphical model of generative attention modeling is illustrated in Figure \ref{graphicModel} with solid and dashed lines.

In our framework, the CVAE cannot be directly and solely optimized due to the unavailability of ground truth $\lambda_t$.
Therefore, we propose to train attention module and CVAE in an alternating way, \ie, we first update CVAE with ``pseudo label'' of $\lambda_t$ given by the attention module, and then train attention module with fixed CVAE.
The two stages are repeated for several iterations.
Since there exist other loss terms for attention modeling (\eg $\mathcal{L}_d$), the pseudo label can be high-quality and hence a good convergence can be reached.
Experimental results empirically validate it.

\subsection{Optimization}

In addition to the above objectives, we exploit a self-guided regularization~\cite{nguyen2019weakly} to further refine the attention.
The temporal class activation maps (TCAM)~\cite{nguyen2018weakly,zhou2016learning} are utilized to produce the \emph{top-down, class-aware} attention maps.
In specific, given a video with label $y$, the TCAM are computed by
\begin{align}\small
    \hat{\lambda}_t^{fg} &= G(\sigma_{s}) \ast \frac{\exp^{\mathbf{w}_y^{T} \mathbf{x}_t}}{\sum_{c = 0}^{C} \exp^{\mathbf{w}_c^{T} \mathbf{x}_t}}, \\
    \hat{\lambda}_t^{bg} &= G(\sigma_{s}) \ast \frac{\sum_{c = 1}^{C} \exp^{\mathbf{w}_c^{T} \mathbf{x}_t}}{\sum_{c = 0}^{C} \exp^{\mathbf{w}_c^{T} \mathbf{x}_t}},
\end{align}
where $\mathbf{w}_c$ indicates the parameters of the classification module for class $c$. $\hat{\lambda}_t^{fg}$ and $\hat{\lambda}_t^{bg}$ are foreground and background TCAM, respectively.
$G(\sigma_{s})$ is a Gaussian smooth filter with standard deviation $\sigma_{s}$, and $\ast$ represents convolution.
The generated $\hat{\lambda}_t^{fg}$ and $\hat{\lambda}_t^{bg}$ are expected to be consistent with the \emph{bottom-up, class-agnostic} attention $\bm{\lambda}$, hence the loss $\mathcal{L}_{guide}$ can be formulated as
\begin{equation}\small
    \mathcal{L}_{guide} = \frac{1}{T} \sum_{t = 1}^{T} |\lambda_t - \hat{\lambda}_t^{fg}| + |\lambda_t - \hat{\lambda}_t^{bg}|.
\end{equation}

To sum up, we optimize the whole framework by alternately executing the following two steps:
\begin{enumerate}
    \item Update attention and classification modules with loss
    \begin{equation}\small
    \label{L}
        \mathcal{L} = \mathcal{L}_d + \gamma_1 \mathcal{L}_{re} + \gamma_2 \mathcal{L}_{guide},
    \end{equation}
    where $\gamma_1,\gamma_2$ denote the hyper-parameters.
    \item Update CVAE with loss $\mathcal{L}_{CVAE}$.
\end{enumerate}
The whole architecture is illustrated in Figure~\ref{overall architecture}.

\subsection{Action Prediction}

To generate action proposals for a video during inference, we feed the video to DGAM and obtain the attention $\bm{\lambda} = (\lambda_t)_{t = 1}^{T}$.
By filtering out frames with attention lower than a threshold $t_{att}$, we extract consecutive segments with high attention values as the predicted locations.
For each segment $[t_s, t_e]$, we temporally pool the features with attention, and get the classification score $s(t_s, t_e, c)$ for class $c$, which is the output of classification module before softmax.
We further follow \cite{shou2018autoloc, liu2019completeness} to refine $s(t_s, t_e, c)$ by subtracting the score of its surroundings. The final score $s^\ast(t_s, t_e, c)$ is calculated by
\begin{equation}\small
\begin{split}
    s^\ast (t_s, t_e, c) &= s(t_s, t_e, c) - \eta\cdot s(t_s - \frac{t_e - t_s}{4}, t_s, c) \\&\quad - \eta\cdot s(t_e, t_e + \frac{t_e - t_s}{4}, c),
\end{split}
\end{equation}
where $\eta$ is the subtraction parameter.

\section{Experiments}

\subsection{Datasets and Evaluation Metrics}

For evaluation, we conduct experiments on two benchmarks, THUMOS14~\cite{idrees2017thumos} and ActivityNet1.2~\cite{caba2015activitynet}. During training, only video-level category labels are available.

\textbf{THUMOS14} contains videos from 20 classes for action localization task.
We follow the convention to train on validation set with 200 videos and evaluate on test set with 212 videos.
Note that we exclude the wrongly annotated video\#270 from test set, following~\cite{nguyen2019weakly,zhao2017temporal}.
This dataset is challenging for its finely annotated action instances.
Each video contains 15.5 action clips on average.
Length of action instance varies widely, from a few seconds to minutes.
Video length also ranges from a few seconds to 26 minutes, with an average of around 3 minutes.
Compared to other large-scale datasets, \eg, ActivityNet1.2, THUMOS14 has less training data which indicates higher requirement of model's generalization ability and robustness.

\textbf{ActivtyNet1.2} contains 100 classes of videos with both video-level labels and temporal annotations.
Each video contains 1.5 action instances on average. Following \cite{wang2017untrimmednets,shou2018autoloc}, we train our model on training set with 4819 videos and evaluate on validation set with 2383 videos.

\textbf{Evaluation Metrics.} We follow the standard evaluation protocol and report mean Average Precision (mAP) at different intersection over union (IoU) thresholds.
The results are calculated using the benchmark code provided by ActivityNet official codebase\footnote{https://github.com/activitynet/ActivityNet/tree/master/Evaluation}. For fair comparison, all results on THUMOS14 are averaged over five runs.

\begin{table}[t]\small
\caption{Attention evaluation on THUMOS14. The ``Old'' model (O) is trained without the generative attention modeling, and the ``New'' model (N) is our DGAM. We assemble specific models by alternately choosing Attention (Att) and Classification (Cls) modules from the two models.}
\label{table:module_assemble}
\begin{center}
\begin{tabular}{c c|c c c c c}
\hline
Att & Cls & \multicolumn{5}{c}{mAP@IoU} \\
 & & 0.3 & 0.4 & 0.5 & 0.6 & 0.7 \\
\hline \hline
O & O & 43.8 & 35.8 & 26.7 & 18.2 & 9.7 \\
O & N & 44.2 & 36.1 & 27.0 & 18.7 & 9.8 \\
N & O & 46.1 & 38.2 & 28.8 & 19.4 & 11.2 \\
N & N & 46.8 & 38.2 & 28.8 & 19.8 & 11.4 \\
\hline
\end{tabular}
\end{center}
\vspace{-1em}
\end{table}

\subsection{Implementation Details}

We utilize I3D~\cite{carreira2017quo} network pre-trained on Kinetics~\cite{kay2017kinetics} as the feature extractor\footnote{https://github.com/deepmind/kinetics-i3d}.
In specific, we first extract optical flow from RGB data using TV-L1 algorithm~\cite{perez2013tv}.
Then we divide both streams into non-overlapping 16-frame snippets and send them into the pre-trained I3D network to obtain two 1024-dimension feature frames for each snippet.
We train separate DGAMs for RGB and flow streams. The proposals from them are combined with Non-Maximum Suppression (NMS) during inference.
Following~\cite{nguyen2018weakly,nguyen2019weakly}, we set $T$ to 400 for all videos during training.
During evaluation, we feed all frames of each video to our network if the frame number is less than $T_{max}$, otherwise we sample $T_{max}$ frames uniformly.
$T_{max}$ is 400 for THUMOS14, and 200 for ActivityNet1.2.

We set $\alpha = 0.03$ in Eq.~(\ref{L_d}) and $\beta = 0.1$ in Eq.~(\ref{L_CVAE}). In Eq.~(\ref{L}), we set $\gamma_1$ to $0.5$ for RGB stream, and $0.3$ for flow stream. $\gamma_2$ is set as $0.1$.
The whole architecture is implemented with PyTorch~\cite{paszke2017automatic} and trained on single NVIDIA Tesla M40 GPU using Adam optimizer~\cite{kingma2014adam} with learning rate of $10^{-3}$.
To stabilize the training of DGAM, we leverage a warm-up strategy in the first 300 epochs when updating $\mathcal{L}_{CVAE}$ and $\mathcal{L}_{re}$.

\begin{table}[t]\small
\caption{Statistics comparison on THUMOS14 with/without generative attention modeling. $\downarrow$ indicates lower is better, $\uparrow$ indicates higher is better. For details of notation, please refer to Section \ref{sec:attention_eval}.}
\label{table:statistics}
\begin{center}
\begin{tabular}{c r|c c}
\hline
Metric & & w/o & w/ \\
\hline \hline
$|att - gt| \ / \ |gt|$ & $\downarrow$ & 0.777 & \textbf{0.698} \\
\hline
$|gt - att| \ / \ |gt|$ & $\downarrow$ & 0.858 & \textbf{0.707} \\
\hline
$|(cls - gt) \cap \overline{att}| \ / \ |gt|$ & $\uparrow$ & 1.522 & \textbf{1.543} \\
\hline
$|(att \cap gt) - cls| \ / \ |gt|$ & $\uparrow$ & \textbf{0.001} & \textbf{0.001}\\
\hline
\end{tabular}
\end{center}
\vspace{-1em}
\end{table}

\begin{table*}[h!] \small
\caption{Results on THUMOS14 testing set. We report mAP values at IoU thresholds 0.1:0.1:0.9. Recent works in both fully-supervised and weakly-supervised settings are reported. UNT and I3D represent UntrimmedNet and I3D feature extractor, respectively. Our method outperforms the state-of-the-art methods, especially at high IoU threshold, which means that our model could produce finer and more precise predictions.
Compared to fully-supervised methods, our DGAM can achieve close or even better performance.
}
\label{table1}
\vspace{-0.05in}
\begin{center}
\begin{tabular}{c|c|c||c c c c c c c c c}
\hline
Method & Supervision & Feature & \multicolumn{9}{c}{mAP@IoU} \\
 & & & 0.1 & 0.2 & 0.3 & 0.4 & 0.5 & 0.6 & 0.7 & 0.8 & 0.9\\
\hline\hline
S-CNN~\cite{shou2016temporal} & Full & - & 47.7 & 43.5 & 36.3 & 28.7 & 19.0 & 10.3 & 5.3 & - & - \\
R-C3D~\cite{xu2017r} & Full & - & 54.5 & 51.5 & 44.8 & 35.6 & 28.9 & - & - & - & - \\
SSN~\cite{zhao2017temporal} & Full & - & 66.0 & 59.4 & 51.9 & 41.0 & 29.8 & - & - & - & - \\
Chao \etal~\cite{chao2018rethinking} & Full & - & 59.8 & 57.1 & 53.2 & 48.5 & 42.8 & 33.8 & 20.8 & - & -\\
BSN~\cite{lin2018bsn} & Full & - & - & - & 53.5 & 45.0 & 36.9 & 28.4 & 20.0 & - & -\\
P-GCN~\cite{zeng2019graph} & Full & - & \textbf{69.5} & \textbf{67.8} & \textbf{63.6} & \textbf{57.8} & \textbf{49.1} & - & - & - & -\\
\hline \hline
Hide-and-Seek~\cite{singh2017hide} & Weak & - & 36.4 & 27.8 & 19.5 & 12.7 & 6.8 & - & - & - & -\\
UntrimmedNet~\cite{wang2017untrimmednets} & Weak & - & 44.4 & 37.7 & 28.2 & 21.1 & 13.7 & - & - & - & -\\
Zhong \etal~\cite{zhong2018step} & Weak & - & 45.8 & 39.0 & 31.1 & 22.5 & 15.9 & - & - & - & -\\
AutoLoc~\cite{shou2018autoloc} & Weak & UNT & - & - & 35.8 & 29.0 & 21.2 & 13.4 & 5.8 & - & -\\
CleanNet~\cite{liuweakly} & Weak & UNT & - & - & 37.0 & 30.9 & 23.9 & 13.9 & 7.1 & - & -\\
\hline
STPN~\cite{nguyen2018weakly} & Weak & I3D & 52.0 & 44.7 & 35.5 & 25.8 & 16.9 & 9.9 & 4.3 & 1.2 & 0.1\\
MAAN~\cite{yuan2018marginalized} & Weak & I3D & 59.8 & 50.8 & 41.1 & 30.6 & 20.3 & 12.0 & 6.9 & 2.6 & 0.2 \\
W-TALC~\cite{paul2018w} & Weak & I3D & 55.2 & 49.6 & 40.1 & 31.1 & 22.8 & - & 7.6 & - & -\\
Liu \etal~\cite{liu2019completeness} & Weak & I3D & 57.4 & 50.8 & 41.2 & 32.1 & 23.1 & 15.0 & 7.0 & - & -\\
TSM~\cite{yutemporal} & Weak & I3D & - & - & 39.5 & - & 24.5 & - & 7.1 & - & -\\
3C-Net~\cite{narayan20193c} & Weak & I3D & 56.8 & 49.8 & 40.9 & 32.3 & 24.6 & - & 7.7 & - & -\\
Nguyen \etal~\cite{nguyen2019weakly} & Weak & I3D & \textbf{60.4} & \textbf{56.0} & 46.6 & 37.5 & 26.8 & 17.6 & 9.0 & 3.3 & \textbf{0.4}\\
DGAM & Weak & I3D & 60.0 & 54.2 & \textbf{46.8} & \textbf{38.2} & \textbf{28.8} & \textbf{19.8} & \textbf{11.4} & \textbf{3.6} & \textbf{0.4}\\
\hline
\end{tabular}
\end{center}
\vspace{-0.25in}
\end{table*}

\subsection{Statistical Evaluation on Attention}
\label{sec:attention_eval}
We first evaluate the learned attention of DGAM and its effectiveness on handling action-context confusion.
For comparison, an ``old'' model is trained by removing the generative attention modeling (GAM) from DGAM, and our DGAM is denoted as the ``new'' model.
Note that only \emph{Attention} and \emph{Classification} modules are involved during inference.
When evaluating, we assemble specific models by alternately choosing the two modules from ``old'' or ``new'' models.
Table \ref{table:module_assemble} details the mAP results on THUMOS14.
It can be found that the new attention module largely improves the performance, while there is little or no improvement with the new classification module.
This observation indicates that DGAM indeed learns better attention values. Even with ``old'' classifier, the ``new'' attention can boost the localization significantly.

We further collect several statistics to show the improvement intuitively in Table \ref{table:statistics}.
Experiments are conducted on both ``old" (w/o GAM) and ``new" (w/ GAM) models.
In particular, \emph{att} (\emph{cls}) indicates the set of frames with attention values (classification scores) larger than a threshold $t^*=0.5$, and \emph{gt} is the set of ground truth frames.
$|\cdot|$ represents size of a set. `$a-b$', `$a \cap b$' and `$\overline{a}$' indicate set exclusion, intersection and complement, separately. Though such simple thresholding is not exactly the predicted locations, it somewhat reflects the quality of localization.

In Table \ref{table:statistics}, $|att-gt|/|gt|$ or $|gt-att|/|gt|$ indicates the percentage of frames falsely captured or omitted by attention.
It shows that both false activation and omission can be reduced with GAM.
Moreover, an improvement in $|(cls-gt) \cap \overline{att}| / |gt|$ demonstrates that GAM can better filter out the false positives (\eg context frames) made by classifier.
$|(att \cap gt) - cls|/|gt|$ measures how attention can capture the false negatives, \ie, action frames neglected by classifier.
Since GAM is devised for excluding the false positives produced by classifier, it is not surprising that GAM contributes little to it.

\begin{table}[t]
\caption{Contribution of each design in DGAM on THUMOS14. Note that when adding $\mathcal{L}_{re}$, $\mathcal{L}_{CVAE}$ is involved simultaneously.}
\label{table:ablation}
\begin{center}
\begin{tabular}{c c c c|c}
\hline
$\mathcal{L}_{fg}$ & $\mathcal{L}_{bg}$ & $\mathcal{L}_{guide}$ & $\mathcal{L}_{re}$ & mAP@0.5 \\
\hline \hline
\checkmark & - & - & - & 21.5 \\
\checkmark & \checkmark & - & - & 24.8 \\
\checkmark & \checkmark & \checkmark & - & 26.7 \\
\checkmark & \checkmark & \checkmark & \checkmark & 28.8 \\
\hline
\end{tabular}
\end{center}
\vspace{-1.8em}
\end{table}

\subsection{Ablation Studies}

Next we study how each component in DGAM influences the overall performance.
We start with the basic model that directly optimizes the attention based foreground classification loss $\mathcal{L}_{fg}$. The background classification loss $\mathcal{L}_{bg}$, the self-guided regularization loss $\mathcal{L}_{guide}$, and the feature reconstruction loss $\mathcal{L}_{re}$ are further included step by step. Note that adding $\mathcal{L}_{re}$ indicates involving the generative attention modeling, where $\mathcal{L}_{CVAE}$ is also optimized.

Table \ref{table:ablation} summarizes the performance by considering one more factor at each stage on THUMOS14.
Background classification is a general approach for both video recognition and localization. In our case, it is part of our discriminative attention modeling, which brings a performance gain of 3.3\%.
Self-guided regularization is the additional optimization of our system, which leads to 1.9\% mAP improvement.
Our generative attention modeling further contributes a significant increase of 2.1\% and the performance of DGAM finally reaches 28.8\%.

\begin{table*}[h!]\small
\caption{Results on ActivityNet1.2 validation set.
We report mAP at different IoU thresholds and mAP@AVG (average mAP on thresholds 0.5:0.05:0.95).
Note that $^\ast$ indicates utilization of weaker feature extractor than others.
Our method outperforms state-of-the-art methods by a large margin, where an improvement of 2\% is made on mAP@AVG.
Our result is also comparable to fully-supervised models.}
\label{table2}
\begin{center}
\begin{tabular}{c|c||c c c c c c c c c c|c}
\hline
Method & Supervision & \multicolumn{11}{c}{mAP@IoU} \\
 & & 0.5 & 0.55 & 0.6 & 0.65 & 0.7 & 0.75 & 0.8 & 0.85 & 0.9 & 0.95 & AVG\\
\hline\hline
SSN~\cite{zhao2017temporal} & Full & 41.3 & 38.8 & 35.9 & 32.9 & 30.4 & 27.0 & 22.2 & 18.2 & 13.2 & 6.1 & 26.6 \\
\hline \hline
UntrimmedNet$^\ast$~\cite{wang2017untrimmednets} & Weak & 7.4 & 6.1 & 5.2 & 4.5 & 3.9 & 3.2 & 2.5 & 1.8 & 1.2 & 0.7 & 3.6\\
AutoLoc$^\ast$~\cite{shou2018autoloc} & Weak & 27.3 & 24.9 & 22.5 & 19.9 & 17.5 & 15.1 & 13.0 & 10.0 & 6.8 & 3.3 & 16.0\\
\hline
W-TALC~\cite{paul2018w} & Weak & 37.0 & 33.5 & 30.4 & 25.7 & 14.6 & 12.7 & 10.0 & 7.0 & 4.2 & 1.5 & 18.0\\
TSM~\cite{yutemporal} & Weak & 28.3 & 26.0 & 23.6 & 21.2 & 18.9 & 17.0 & 14.0 & 11.1 & 7.5 & 3.5 & 17.1\\
3C-Net~\cite{narayan20193c} & Weak & 35.4 & - & - & - & 22.9 & - & - & - & 8.5 & - & 21.1\\
CleanNet~\cite{liuweakly} & Weak & 37.1 & 33.4 & 29.9 & 26.7 & 23.4 & 20.3 & 17.2 & 13.9 & 9.2 & 5.0 & 21.6\\
Liu \etal~\cite{liu2019completeness} & Weak & 36.8 & - & - & - & - & 22.0 & - & - & - & \textbf{5.6} & 22.4\\
DGAM & Weak & \textbf{41.0} & \textbf{37.5} & \textbf{33.5} & \textbf{30.1} & \textbf{26.9} & \textbf{23.5} & \textbf{19.8} & \textbf{15.5} & \textbf{10.8} & 5.3 & \textbf{24.4}\\
\hline
\end{tabular}
\end{center}
\vspace{-0.15in}
\end{table*}

\begin{figure}[h]
     \centering
     \includegraphics[width=0.85\linewidth]{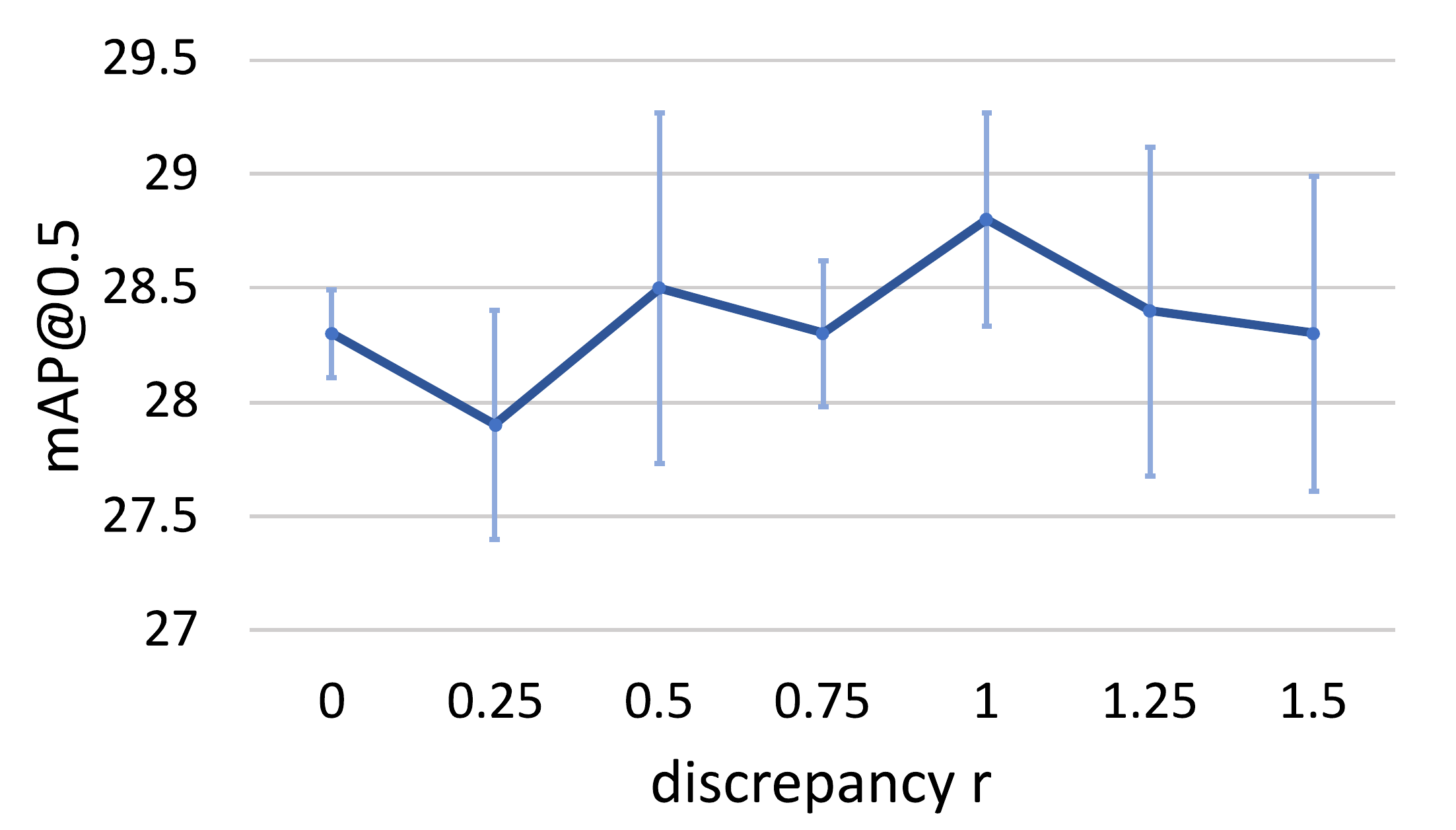}
     \caption{Evaluation on latent prior discrepancy $r$ on THUMOS14. We show mAP@$0.5$ with different $r$. Larger $r$ indicates larger discrepancy between priors of $\mathbf{z}_t$ under different attentions $\lambda_t$.}
     \label{prior_discrepancy}

\end{figure}

\begin{table}[h!]\small
\caption{Evaluation on dimension of latent space on THUMOS14. We experiment with different dimensions of $2^n$, $n = 4, 5, \cdots, 9$.}
\label{latent_dimension}
\vspace{-2em}
\begin{center}
\begin{tabular}{c|c c c c c c}
\hline
$\log_2$(dim)& 4 & 5 & 6 & 7 & 8 & 9\\
\hline \hline
mAP@0.5 & 26.5 & 27.5 & 28.0 & 28.8 & 28.3 & 27.7\\
\hline
\end{tabular}
\end{center}

\end{table}

\begin{table}[h!]\small
\caption{Evaluation on parameter for reconstruction-sampling trade-off in CVAE. mAP@$0.5$ is reported on THUMOS14.}
\label{trade_off_CVAE}
\vspace{-2em}
\begin{center}
\begin{tabular}{c|c c c c c c c c}
\hline
$\beta$ & 0.01 & 0.03 & 0.07 & 0.1 & 0.3 & 0.7\\
\hline \hline
mAP@0.5 & 28.2 & 28.1 & 28.4 & 28.8 & 28.0 & 28.4\\
\hline
\end{tabular}
\end{center}

\end{table}

\subsection{Evaluation on Parameters}

To further understand the proposed model, we conduct evaluations to analyze the impact of different parameter settings in DGAM. mAP@$0.5$ on THUMOS14 is reported.

\textbf{Discrepancy between latent prior of different $\lambda_t$}.
In generative attention modeling, different attentions $\lambda_t$ correspond to different feature distributions $p_\psi (\mathbf{x}_t | \lambda_t)$.
The discrepancy between these distributions can be implicitly modeled by the
discrepancy between latent codes $\mathbf{z}_t$ sampled from different priors, which are modeled as different Gaussian distributions $p_\psi (\mathbf{z}_t | \lambda_t)=\mathcal{N}(\mathbf{z}_t|r \lambda_t \cdot \mathbf{1},~\mathbf{I})$.
Here $r$ controls the discrepancy.
We evaluate $r$ every 0.25 from 0 to 1.5, and the results are shown in Figure~\ref{prior_discrepancy}.
In general, the performance is relatively stable with small fluctuation, demonstrating the robustness of DGAM.

\textbf{Dimension of latent space}.
The dimension of latent space in CVAE is crucial for quality of reconstruction and complexity of modeled distribution.
High dimension can facilitate the approximation of feature distribution, hence leading to more accurate attention learning.
However, more training data is also required.
We evaluate different dimensions of $2^n$, $n = 4, 5, \cdots, 9$. As shown in Table~\ref{latent_dimension}, mAP improves rapidly with increasing dimension, which indicates better generative attention modeling.
The result reaches the peak at dimension $2^7 = 128$. After that, the performance starts dropping, partially because of the sparsity of limited data in high-dimensional latent space.

\textbf{Reconstruction-sampling trade-off in CVAE}. The hyper-parameter $\beta$ in Eq.~(\ref{L_CVAE}) balances reconstruction quality (the first term) and sampling accuracy (the second term).
With larger $\beta$, we expect the approximated posterior to be closer to the prior, which improves the precision when sampling latent vectors from prior, while the reconstruction quality (\ie the quality of learned distribution) will decrease.
We test different $\beta$ from 0 to 1. As shown in Table~\ref{trade_off_CVAE}, the performance fluctuates in a small range from 28\% to 28.8\%, indicating that our method is insensitive to $\beta$.

\subsection{Comparisons with State-of-the-Art}

Table~\ref{table1} compares our DGAM with existing approaches in both weakly-supervised and fully-supervised action localization on THUMOS14.
Our method outperforms other weakly-supervised methods, especially at high IoU threshold, which means DGAM could produce finer and more precise predictions.
Compared with state of the art, DGAM improves mAP at IoU=0.5 by 2\%.
Note that Nguyen \etal~\cite{nguyen2019weakly} achieves better performance at IoU=0.1 and 0.2 than our model, partially because our generative attention modeling may discard out-of-distribution hard candidates (outliers), which become common when IoU is low.
Furthermore, our results are comparable with several fully-supervised methods, indicating the effectiveness of the proposed DGAM.

On ActivityNet1.2, we summarize the performance comparisons in Table~\ref{table2}.
Our method significantly outperforms the state-of-the-arts.
Particularly, DGAM surpasses the best competitor by 2\% on mAP@AVG.
Our method also demonstrates comparable results to fully-supervised methods.

\section{Conclusion}

We have presented a novel Discriminative and Generative Attention Modeling (DGAM) method to solve the action-context confusion issue in weakly-supervised action localization.
Particularly, we study the problem of modeling frame-wise attention based on the distribution of frame features.
With the observation that context feature obviously differs from action feature, we devise a conditional variation auto-encoder (CVAE) to construct different feature distributions conditioned on different attentions.
The learned CVAE in turn refines the desired frame-wise attention according to their features.
Experiments conducted on two benchmarks, \ie, THUMOS14 and ActivityNet1.2, validate our method and analysis. More remarkably, we achieve the new state-of-the-art results on both datasets.

{\small \textbf{Acknowledgements}~ This work is supported by Beijing Municipal Commission of Science and Technology under Grant Z181100008918005, National Natural Science Foundation of China (NSFC) under Grant 61772037. Baifeng Shi thanks Prof. Tingting Jiang and Daochang Liu for enlightening discussions.}

{\small
\bibliographystyle{ieee_fullname}
\bibliography{egbib}
}

\end{document}